\documentclass[10pt]{article} 
\usepackage[preprint]{tmlr}

\usepackage{amsmath,amsfonts,bm}

\def\eqref#1{equation~\ref{#1}}

\def\1{\bm{1}}

\DeclareMathAlphabet{\mathsfit}{\encodingdefault}{\sfdefault}{m}{sl}
\SetMathAlphabet{\mathsfit}{bold}{\encodingdefault}{\sfdefault}{bx}{n}

\usepackage[dvipsnames]{xcolor}
\usepackage{hyperref}
\usepackage{url}
\usepackage{graphicx}
\usepackage{booktabs}
\usepackage{multirow}
\usepackage{float}
\usepackage{listings}
\usepackage{adjustbox}
\usepackage{subcaption}
\usepackage{cleveref}
\title{Contrastive Joint‑Embedding Prediction \\for Representation Learning in Structural MRI}

\author{\name Fabian Mager \email fmager@dtu.dk \\
      \addr  Department of Applied Mathematics and Computer Science\\
      Technical University of Denmark
      \AND
      \name Lars Kai Hansen \email lkai@dtu.dk \\
      \addr Department of Applied Mathematics and Computer Science\\
      Technical University of Denmark
}

\def\month{MM}  
\def\year{YYYY} 
\def\openreview{\url{https://openreview.net/forum?id=XXXX}} 

\usepackage{xspace}
\newcommand{\avg}{{\normalsize\textsc{avg}}\xspace}
\newcommand{\cls}{{\normalsize\textsc{cls}}\xspace}
\newcommand{\attn}{{\normalsize\textsc{attn}}\xspace}
\newcommand{\x}{\mathbf{x}}

\newcommand{\zctxt}{z^{\mathrm{ctxt}}}
\newcommand{\ztrgt}{z^{\mathrm{trgt}}}
\newcommand{\ztrgthat}{\hat{z}^{\mathrm{trgt}}}
\newcommand{\zglob}{z^{\mathrm{glob}}}
\newcommand{\mctxt}{m^{\mathrm{ctxt}}}
\newcommand{\mtrgt}{m^{\mathrm{trgt}}}
\newcommand{\X}{\mathbf{X}}
\newcommand{\q}{\mathbf{q}}

\newcommand{\Lcontr}{\mathcal{L}_{\textsc{contr}}}
\newcommand{\Lpred}{\mathcal{L}_{\textsc{pred}}}
\newcommand{\lambdacontr}{\lambda_{\textsc{contr}}}
\newcommand{\lambdapred}{\lambda_{\textsc{pred}}}
\newcommand{\ssize}{\scriptsize}
\newcommand{\spm}[1]{{\ssize$\pm$#1}}
\newcommand{\scls}{{\ssize\textsc{cls}}}
\newcommand{\savg}{{\ssize\textsc{avg}}}
\newcommand{\sattn}{{\ssize\textsc{attn}}}

\newcommand{\ie}{i.e.\@\xspace}

\begin{document}

\maketitle

\begin{abstract}
	Self-supervised learning offers a compelling approach for medical imaging, where labeled data are scarce and acquisition costs are high. We present COJEPA, a self-supervised framework for volumetric brain MRI that combines a joint-embedding predictive architecture (JEPA) with a contrastive loss (CO), targeting two complementary properties: local predictivity and global discriminability. The model is trained without labels on T1-weighted structural MRI from two cohorts (HCP-YA and AABC, $N{=}2286$, ages 22 to 90), extending I-JEPA~\citep{JEPA} to 3D with foreground-aware block masking, a hierarchical convolutional patch embedding, and world-space sinusoidal positional encodings. We evaluate all three objectives across zero-shot twin retrieval, brain tumor segmentation (BraTS 2024), and age regression (OpenBHB). COJEPA achieves the best monozygotic twin recall at rank~1 (0.84), the best finetuning age MAE (2.55\,yr on OpenBHB 3.0T), and matches CO on BraTS whole-tumor Dice, demonstrating that the combined objective yields representations that are simultaneously discriminative and locally structured. Code is available at \url{XXX}.
\end{abstract}

\paragraph{Self-Supervised Learning.}
Self-supervised learning (SSL) is a machine learning paradigm in which the model is trained without the need for labeled data. It leverages the data's inherent structure to generate labels, making it possible to learn representations without the need for manually labeled datasets. Unlabeled data are widely available, reducing dependencies on extensive and costly data labeling. This is especially relevant in the biomedical domain, as large-scale clinical trials are extremely costly if not infeasible to obtain for some patient groups.

\paragraph{SSL in medical imaging.}
In computer vision, SSL is capable of learning meaningful representations, which generalize well and can be finetuned for a variety of downstream tasks \citep{JEPA, DATA2VEC, BEIT, DINO, DINOv2, MOCO}. Despite rapid progress in the domain of natural images and video, SSL has not yet achieved a comparable impact in the medical domain, \ie 3D structural magnetic resonance imaging (MRI). One constraint is the scarcity of large, diverse, and shareable datasets due to acquisition cost, time, privacy, and ethical restrictions. In the case of magnetic resonance imaging, the heterogeneity between scanners, sites, field strengths, and protocols further complicates the pooling of datasets \citep{DINSDALE_2021, KUSHOL_2023}.
Another constraint is the computational burden. Many SSL recipes benefit from large batch sizes, heavy augmentations, and long schedules. For 3D MRI, high spatial resolution and volumetric memory footprints make these setups impractical. Nevertheless, recent works like \citet{BRAINIAC}, a SimCLR-like model \citep{SIMCL} trained on $\approx$50{,}000 brain MRIs, show excellent downstream performance on a variety of tasks.

\paragraph{Brain MRI foundation models.}
Recent brain MRI foundation models span three pretraining paradigms. Volumetric MAE models such as AMAES \citep{Munk2024AMAESAM}, BrainSegFounder \citep{Cox2024BrainSegFounderT3}, and BM-MAE \citep{Robinet2025MultimodalMA} encode full 3D context but incur reconstruction costs that scale cubically with resolution. Contrastive approaches such as BrainIAC \citep{BRAINIAC} and \citet{Kaczmarek2025BuildingAG} demonstrate strong label efficiency via volumetric SimCLR training on large-scale clinical data. Self-distillation methods such as BrainDINO \citep{wu2026braindino} and BrainFound \citep{Mazher2025TowardsGF} operate on 2D slices to reach otherwise prohibitive training scales, at the cost of volumetric context. The recent FOMO25 challenge \citep{munk2026towards}, benchmarking 19 brain MRI foundation models across clinical tasks, confirms no single paradigm consistently dominates: MAE favors segmentation while contrastive objectives favor classification. To our knowledge, no existing brain MRI foundation model adopts joint-embedding predictive pretraining.

\paragraph{Contrastive vs. Masked Image Modelling.}
SSL in vision largely revolves around two paradigms: (i) \emph{Contrastive Learning} (CL) and (ii) \emph{Masked Image Modelling} (MIM). CL methods such as \citep{SIMCL, BYOL, SWAV, MOCO} align the embeddings for augmented views of the same image while separating the embeddings of different images, producing globally discriminative and view-invariant features. In contrast, MIM trains a model to predict the content of masked regions and has proven effective for transformers. Early methods such as \citep{MAE, BEIT, SIMIM} reconstruct in \emph{pixel space}, while more recent approaches (e.g., \citep{JEPA, DATA2VEC, DINO, DINOv2}) predict targets in \emph{token/representation space}. This shift is associated with stronger semantics while avoiding unnecessary pixel decoding \citep{DINOv2,MAE}.

\paragraph{Design principles: predictivity and specificity.}
For 3D brain structure, we posit that useful representations should be (i) \emph{locally predictive} and (ii) \emph{globally specific}. Local predictivity denotes the ability to anticipate distinct anatomical content from context (capturing statistical dependencies of local structure), while global specificity denotes subject-level discriminability and the retention of unique identifiable features. Viewed through this lens, MIM-style objectives naturally promote predictivity, whereas contrastive objectives promote specificity by enforcing global separability among subjects.

\paragraph{Combining contrastive \& predictive objectives.}
Pure CL faces several limitations when applied to 3D MRI. First, CL objectives typically require very large batch sizes to provide sufficient negative examples, which is computationally infeasible for high‑resolution volumetric data. Second, the types of invariances used in natural images do not necessarily translate well to medical images. Joint-embedding prediction (e.g., I-JEPA) instead learns by predicting latent representations of masked regions from a contextual embedding, without pixel-space decoders \citep{JEPA}. However, joint-embedding objectives can be susceptible to representation collapse, as the objective fails if the target embedding becomes constant. \citet{MOMA} show that MIM benefits from a contrastive objective, learning reconstructions that resemble better global semantics. We make a similar claim for joint-embedding predictive architectures: incorporating a contrastive term injects an explicit discriminative bias that preserves semantic diversity and stabilizes training.

\paragraph{Contribution.}
We propose a fully self-supervised model of human brain structure trained on T1-weighted MRIs from the Human Connectome Project Young Adult cohort \citep{HCP-YA}. Our method pairs a joint-embedding prediction objective with a contrastive loss, explicitly targeting the principles of predictivity and specificity. We show that our model learns efficiently on a relatively small pretraining dataset. We comprehensively assess the learned representation via (i) statistical exploration, (ii) reconstruction and segmentation, (iii) zero-shot retrieval, and (iv) linear probing.

\section{Background}

We investigate three self-supervised training objectives: (i) Contrastive learning (CO) trains a single encoder on two augmented views of the same subject and minimizes their embedding distance; (ii) a Joint Embedding Predictive Architecture (JEPA) predicts target patch embeddings from a context view, using an EMA-updated target encoder; and (iii) the Contrastive Joint Embedding Predictive Architecture (COJEPA) unifies both: a third augmented view drives the contrastive term on a global token, while the predictive term operates on the spatial tokens of the same context–target pair.


\subsection{Contrastive Learning}

The contrastive objective encourages view‑invariance by pulling together representations of matching pairs while simultaneously pushing apart all non‑matching samples in the minibatch \citep{WU_2018, CHEN_2020b}.  Here we use the symmetric Information Noise Contrastive Estimation (InfoNCE) loss function defined by \citet{CLIP}.

Given a minibatch $\{(x_i, y_i)\}_{i=1}^N$ of paired views with representations
$z_x, z_y$ and pairwise cosine similarities $\mathbf{A} \in \mathbb{R}^{N\times N}$ with elements $a_{ij} \in [-1, 1]$, the contrastive loss for $x\!\to\!y$ is
\begin{equation}
	\Lcontr(\tau) = \mathbb{E}\!\left[-\log \frac{\exp(a_{ii}/\tau)}{\sum_{j=1}^N \exp(a_{ij}/\tau)}\right].
	\label{eq:contrloss}
\end{equation}

The negatives $\{a_{ij}\}$ where ${j \neq i}$ correspond to the off‑diagonal entries in row $i$ of $\mathbf{A}$. To obtain the equivalent loss for $y\!\to\!x$, the indices $i$ and $j$ in \cref{eq:contrloss} are swapped. The final symmetric contrastive objective is given by the average of the two directional losses.

\subsection{Joint Embedding Prediction}

In a joint‑embedding predictive architecture, the goal is to predict the latent representation of an unobserved target view given the latent representations of a context view and the target view location \citep{JEPA}. The construction of the context and target views is crucial: the context view needs to capture sufficient global information, while the target should contain rich semantics that cannot be resolved by simple interpolation. The authors in \citep{JEPA} solve this via block masking: For a given image, sample $K$ target blocks with scale $s^{\mathrm{trgt}}$ and aspect ratios $r^{\mathrm{trgt}}$. Then, sample a context block with scale $s^{\mathrm{ctxt}}$, and remove overlapping regions. We denote patch-wise context and target region masks as $\{\mctxt\}$ and $\{\mtrgt\}_k$ respectively. Note that in \citep{JEPA}, $K=4$, $s^{\mathrm{trgt}}\!\in\![0.15\,, 0.2]$, $r^{\mathrm{trgt}}\! \in \![0.75\,, 1.5]$ and $s^{\mathrm{ctxt}}\! \in \![0.85\,, 1.0]$.
Given a set of target and context patch masks $\{\mtrgt\}$ and $\{\mctxt\}$, a context encoder $f_{\theta}:\!\{x\}\odot\{\mctxt\}\! \mapsto\! \{\zctxt\}$, target encoder $f_{\bar\theta}:\! \{x\}\!\mapsto\!\{z\}$ and a predictor $p_{\phi}:\! \{\zctxt,\mtrgt\}\!\mapsto \!\{\ztrgthat\}$, the average joint embedding prediction loss over the patches $(\{\ztrgt\}, \{\ztrgthat\})$ is

\begin{equation}
	\Lpred \;=\;\mathbb{E}\left[D\left(\{\ztrgthat\},\, \mathrm{sg}\!\left[\{z\} \odot \{\mtrgt\}\right]\right)
		\right],
	\label{eq:predloss}
\end{equation}

where $D(\cdot,\cdot)$ denotes a distance function and $\mathrm{sg}[\,\cdot\,]$ a 'stop gradient' operation. The target encoder weights are updated via an exponential moving average of the context encoder weights.

\section{Contrastive Joint Embedding Prediction}
\label{sec:cojepa}

\begin{figure}[htb]
	\centering
	\includegraphics[width=1.0\linewidth]{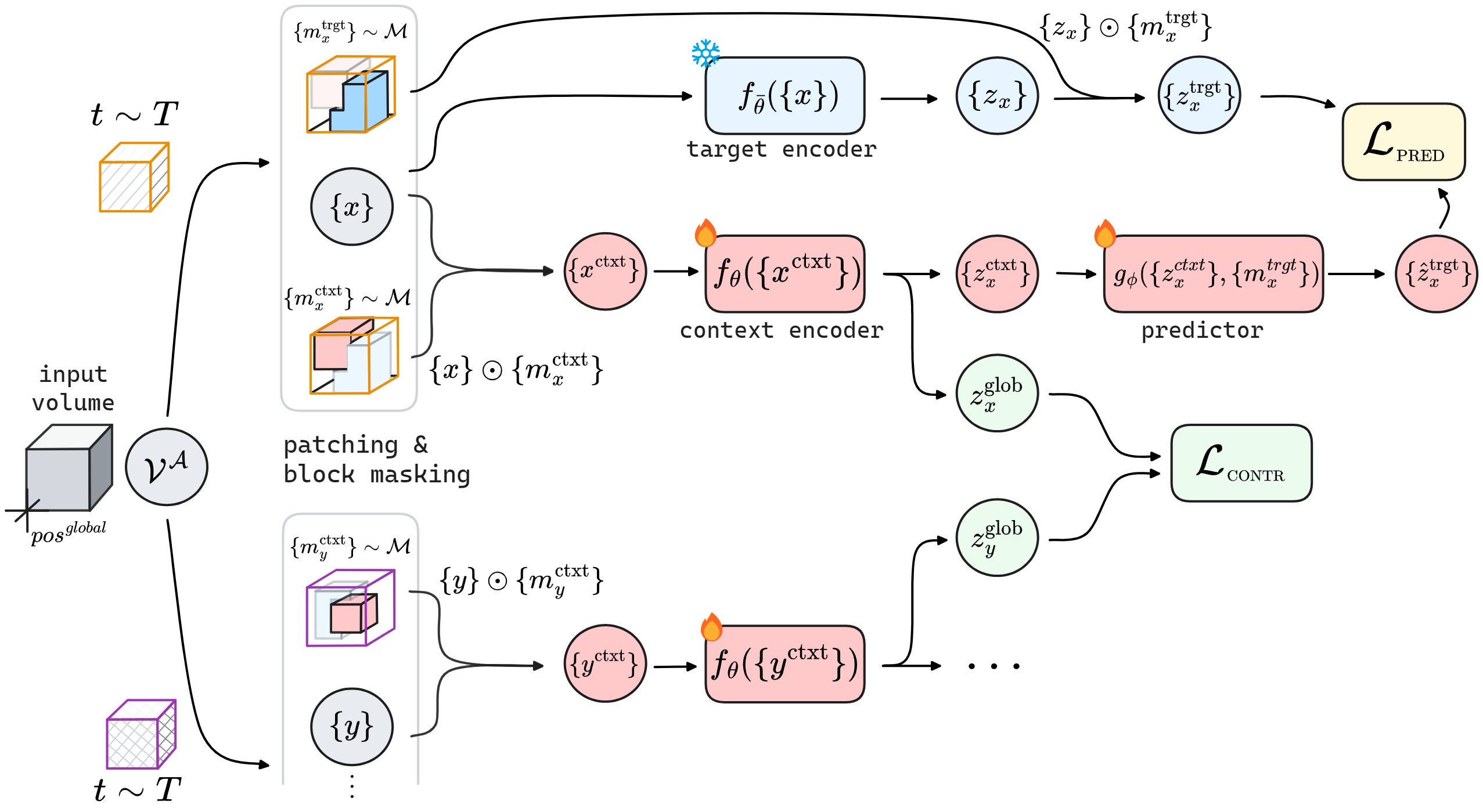}
	\caption{The proposed 3D framework jointly optimizes a predictive objective and a contrastive objective by encoding masked sub-volumes, predicting target embeddings (red/blue), and enforcing global alignment (green), which encourages specificity as well as predictivity of the embeddings. Only the encoder and predictor (red) are trainable. The weights of the offline target encoder (blue) are updated using an exponential moving average of the online encoder weights.}
	\label{fig:pipeline}
\end{figure}

We argue that JEPA combined with a contrastive loss promotes both specificity and predictivity of learned embeddings. The proposed COJEPA architecture introduces subtle modifications compared to the JEPA architecture in \citep{JEPA}: After retrieving patch embeddings and before the transformer layers, we attach an additional token $\zglob$, which then interacts with the spatial context tokens $\zctxt$ via various multi-head attention layers \citep{VASWANI_2017}, and is then used for the contrastive loss.
In addition, we introduce several 3D MRI adaptions to the vision transformer \citep{ViT} architecture used in \citep{JEPA}. An overview of our pipeline is depicted in \cref{fig:pipeline}. All hyperparameters are listed in the supplementary material.

\paragraph{Generating contrastive samples.}
Contrastive learning has been shown to perform best under heavy data augmentations and large batch sizes \citep{SIMCL}. As the contrastive loss merely serves as an auxiliary training objective, we restrict the augmentations to a simple subset: random cropping and scaling, as well as random Gaussian noise and random intensity rescaling.

\paragraph{Positional Encoding.} We take advantage of the fact that the voxels in an MRI exist in a standardized space. Rather than using grid indices of random crops, we feed the model with the global, standardized coordinate of the center voxel. This solves the issue of symmetries in the brain and implicitly encodes the voxel resolution of an input image. Specifically, we assign each patch $P \in \mathbb{R}^{p \times p \times p}$, where $p$ is the patch size, a world space position using the image affine. We then encode the 3D coordinate using a sinusoidal positional embedding with the embedding dimension $D$ split equally across the three spatial axes, resulting in a unique \emph{global positional encoding} $\mathbf{p}_{\textrm{enc}} \in \mathbb{R}^{1\times D}$.

\begin{figure}[tb]
	\centering
	\includegraphics[trim= 50 10 50 10, clip, width=0.9\linewidth]{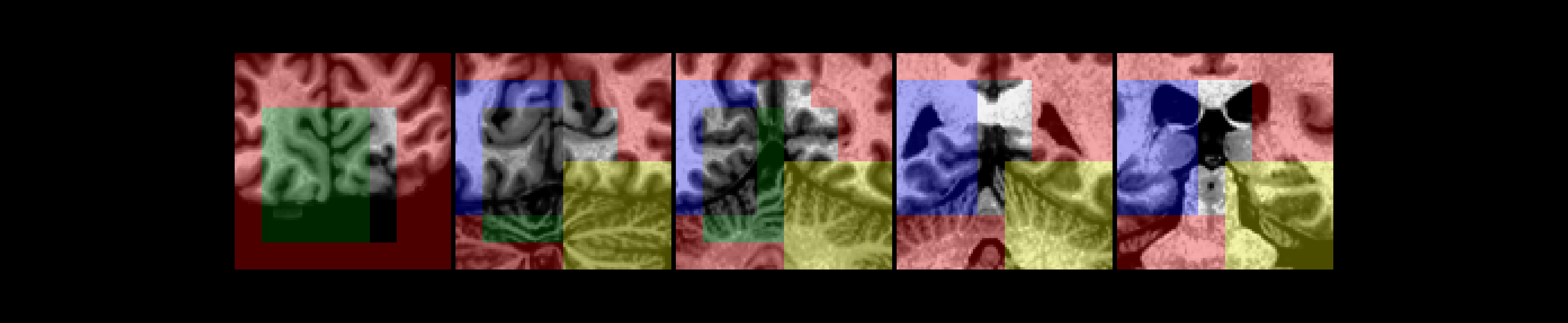}
	\caption{Volume slices with colored mask regions: Context mask $\mctxt$ in red and target masks $\mtrgt$ in blue, green and yellow. All mask regions are non-overlapping. For each sampled target mask, we ensure that the center patch is a foreground patch. All target regions are truncated to the minimum mask size across the batch.}
	\label{fig:masking}
\end{figure}

\paragraph{Masking.} We extend the block masking strategy in \citep{JEPA} to a 3D scenario: First, we sample a random scale $s$ and a random aspect ratio $r$ within the specified limits, which determines the height $h$ and width $w$ of the block mask. We specify the depth by $d = \lfloor h \cdot u + w \cdot (1 - u) \rfloor$, where $u \sim \mathcal{U}(0, 1)$. Following \citep{JEPA}, we first sample $K$ target masks. In order to avoid blocks with little or no information, we prioritize foreground patches in the sampling process. A foreground patch is defined as any patch $p$ for which $\sum_{v \in p} x_v > t$, where $x_v$ is the voxel intensity and $t$ is a predefined threshold. When sampling target masks, we ensure that the center patch is a foreground patch. In case the randomly sampled mask sizes differ for elements in the batch, we first remove background patches before truncating all masks to the minimum mask size across the batch. \cref{fig:masking} shows five equally spaced slices through a sampled volume, with highlighted context, and target mask regions.

\paragraph{Patch Embeddings.} The standard patch embedding layer in Vision Transformers \citep{ViT} projects non-overlapping patches directly into the token embedding space via a single linear projection. We instead use a hierarchical convolutional stem consisting of three sequential 3D convolutions with kernel size 3, which progressively expand the channel dimension from the input to the full embedding dimension $D$, while their combined strides tile the volume into non-overlapping patches. This multi-stage design allows the network to learn local spatial features before projecting into the token space, rather than treating each patch as a flat vector. Each convolution is followed by batch normalization \citep{BatchNorm} and GELU activation. Batch normalization was found to stabilize training on volumetric MRI data.

\paragraph{Loss.}
The total loss combines a prediction term and an optional contrastive term:

\begin{equation}
	\mathcal{L}_{\textsc{tot}} = \lambdapred\,\Lpred + \lambdacontr\,\Lcontr.
\end{equation}

$\Lpred$ is the mean squared error between the predictor output and the target encoder representations. We only calculate $\Lpred$ over foreground tokens. $\Lcontr$ is an InfoNCE loss \citep{CLIP} computed on the \cls token of the two random crops drawn per subject, with the temperature learned as a scalar parameter via $\tau = \exp(\log\tau)$, initialized at $\tau = 0.07$. We set $\lambdapred = 1.0$ and $\lambdacontr = 0.1$.  An ablation of the convolutional stem, global positional encoding and background loss masking is provided in \cref{tab:ablation}.

\section{Experiments}

\paragraph{Pretraining Datasets.}
For pretraining, we used minimally processed structural T1w magnetic resonance imaging data from the Aging Adult Brain Connectome (AABC) \citep{HCP-A} and Human Connectome Project Young Adult \citep{HCP-YA} (HCP-YA). Data have been preprocessed AC–PC orientation alignment, readout distortion correction, cross modal registration, bias-field correction, and brain extraction \citep{hcp_prepr}. We include the cross-sectional data of AABC only ($N{=}1390$, ages 36–90+). The HCP-YA dataset comprises  1113 healthy subjects (ages 22-37). Among the 1113 subjects, 286 are genetically confirmed monozygotic (MZ) twins (143 pairs), including 10 incomplete pairs, resulting in 138 complete MZ twin pairs. In addition, 170 subjects are genetically confirmed duozygotic twins (DZ) (79 pairs), including 12 incomplete pairs, resulting in 79 complete DZ twin pairs. We divide subjects into a training and hold-out set, where one respective twin of both MZ and DZ is assigned to the hold-out set ($N{=}217$). All other subjects are used for training ($N{=}2286$, HCP-YA $\cup$ AABC).

\paragraph{Preprocessing.}
Raw T1w scans undergo a minimal preprocessing pipeline: The foreground is cropped tightly around the brain using the image intensity. After applying the random augmentations described in \cref{sec:cojepa}, each volume is histogram-normalized to $[0, 1]$ using 128 bins. For each subject, we draw two random crops per iteration, providing positive pairs for the contrastive loss. We use MONAI \citep{MONAI} for data loading and augmentation.

\paragraph{Encoder.}
We use a Vision Transformer Base (ViT-B) \citep{ViT} as both the context and target encoder, with an embedding dimension of 768, 12 transformer blocks, 12 attention heads, and an MLP ratio of 4. Volumes are tokenized using a patch size of $9^3$ voxels via the convolutional patch embedding described in \cref{sec:cojepa}. When trained with a contrastive loss, we attach a \cls token. In addition, we add a single register token \citep{registers}.

\paragraph{Predictor.}
A lightweight ViT predictor takes context features together with target positional embeddings and predicts the target representations at masked locations. The predictor has an embedding dimension of 384, 12 transformer blocks, and 12 attention heads, giving it roughly one quarter the parameter count of the encoder.

\paragraph{Optimization.}
The context and target encoders share the same ViT-Base architecture. The target receives no gradient updates and is updated via an exponential moving average (EMA) of the context encoder weights, with the momentum coefficient annealed from 0.996 to 1.0. We optimize with AdamW, using a cosine learning rate schedule with a linear warmup of 50 epochs, a peak learning rate of $10^{-4}$, and a final learning rate of $10^{-5}$. The weight decay is similarly annealed from 0.04 to 0.4 following a cosine schedule. The model is trained for 4000 epochs on 8 AMD MI250x GPUs using bfloat16 mixed precision, with a per-GPU batch size of 12 crops $\times$2 views per GPU.

\subsection{Evaluation}

To effectively assess the pretraining scheme we compare three variants: \textbf{CO} (contrastive only, InfoNCE on the \cls{} token), \textbf{JEPA} (predictive only), and \textbf{COJEPA} (both objectives combined). Unless stated otherwise, encoder weights are frozen and a lightweight task head is trained on top. For segmentation and age regression we follow a label-efficiency protocol, training the probe at 5, 25, 50, and 100\% of available labels, and additionally report end-to-end finetuning at 100\% of labels.

\paragraph{Principal components of feature space.}
We extract patch-level representations from the target encoder and apply a Principal Component analysis (PCA) of foreground tokens. We inspect the energy in the principal components and visualize the first three components by mapping each component to a color channel, similar to \citep{DINOv2}.

\paragraph{Zero shot retrieval.}
We evaluate whether the representations capture family structure in the dataset by measuring retrieval performance on MZ and DZ twin pairs; only one twin per pair was seen during training. We extract a single global token per subject: the \cls{} token (appended to the sequence and updated via the contrastive loss) for CO and COJEPA, and the \avg{} token (global average over all spatial tokens) for JEPA, which has no \cls{} token. We calculate the cosine similarity between all hold-out subjects and rank them accordingly; the positive match is the co-twin. We report recall at ranks 1, 5, and 10 (R@1, R@5, R@10).

\paragraph{Segmentation.}
We train a lightweight decoder on top of frozen encoder features to perform whole-tumor segmentation on BraTS 2024 \citep{BRATS_2024}, an out-of-distribution dataset with varying scanners, sites, and pathological tissue. The decoder follows an MAE-style design \citep{MAE}: a linear projection maps encoder tokens from dimension 768 to a decoder embedding dimension of 512, followed by 4 transformer blocks with 4 attention heads. The decoder is trained with a Dice loss on $N_{\mathrm{train}}{=}1135$ subjects and evaluated on $N_{\mathrm{test}}{=}729$ subjects, following the label-efficiency protocol described above.

\paragraph{Age prediction.}
We train a linear probe on top of frozen encoder features to predict chronological age on the OpenBHB dataset \citep{OPENBHB} ($N_{\mathrm{train}}{=}3227$, $N_{\mathrm{test}}{=}395$). We restrict evaluation to 3.0\,T scans in the external test split ($N{=}339$, ages 6 to 48\, yrs), matching the field strength of the pretraining data and providing a stringent out-of-distribution assessment across multiple sites. We evaluate three pooling strategies applied to the frozen encoder output: the \cls{} token and the \avg{} token (plus a single linear layer), as well as \attn{} pooling. For the \attn{} pooling, let $\X \in \mathbb{R}^{N\times D}$ be the spatial tokens and $\q \in \mathbb{R}^{1\times D}$ a learned query; we compute attention scores $\alpha = \mathrm{softmax}(\X\q^T / \sqrt{D})$ and probe the weighted average $\bar\x = \alpha^T\X$ using a single linear layer.

\section{Results}

\begin{figure}[t]
	\centering
	\begin{minipage}{0.49\linewidth}
		\begin{subfigure}{0.85\linewidth}
			\centering
			\includegraphics[width=1.0\linewidth]{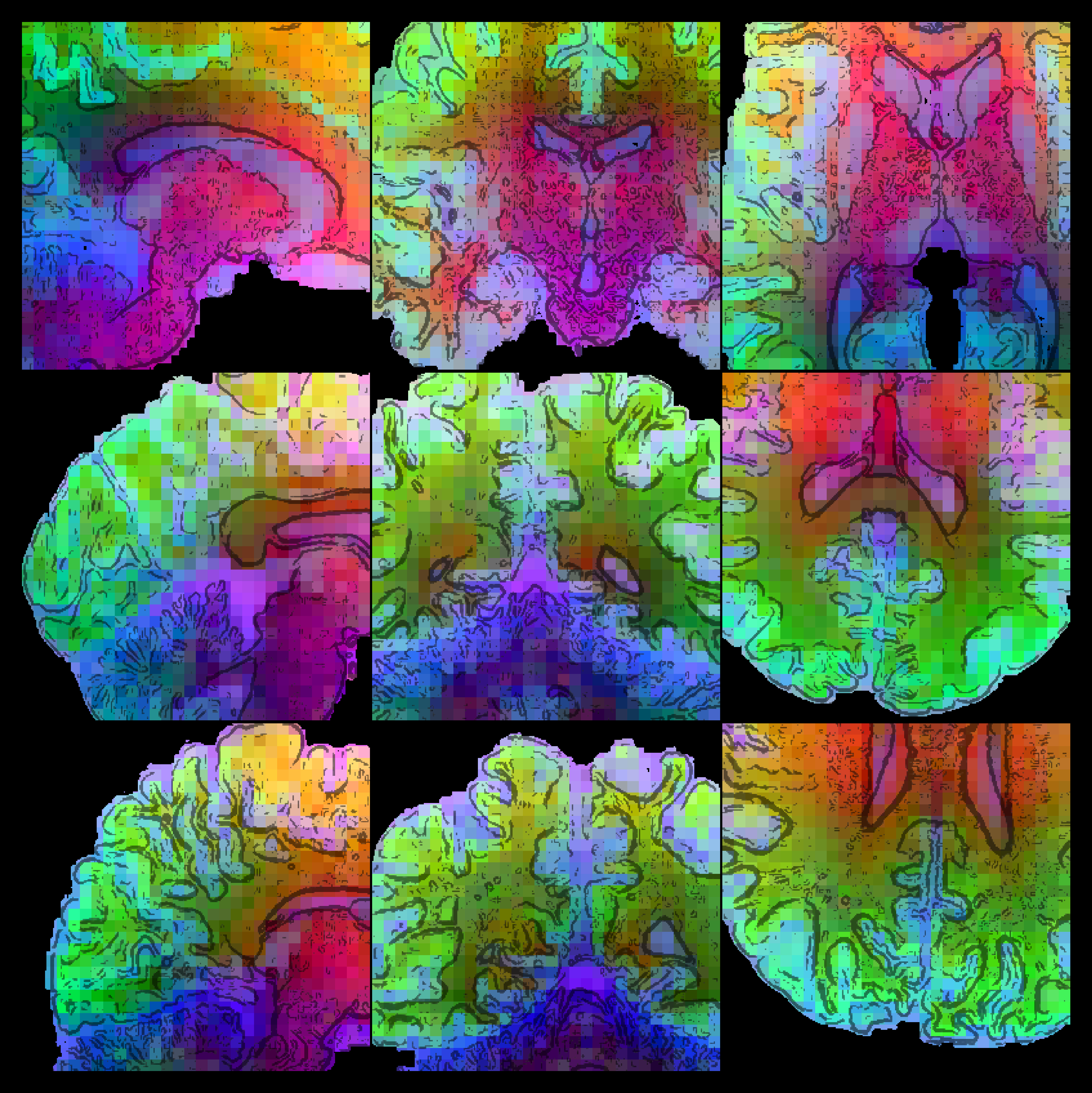}
			\vspace{.5em}
			\caption{PCA visualization of embeddings}
			\label{fig:pca_cojepa}
		\end{subfigure}
	\end{minipage}
	\begin{minipage}{0.49\linewidth}
		\begin{subfigure}{.96\linewidth}
			\centering
			\includegraphics[width=1\linewidth]{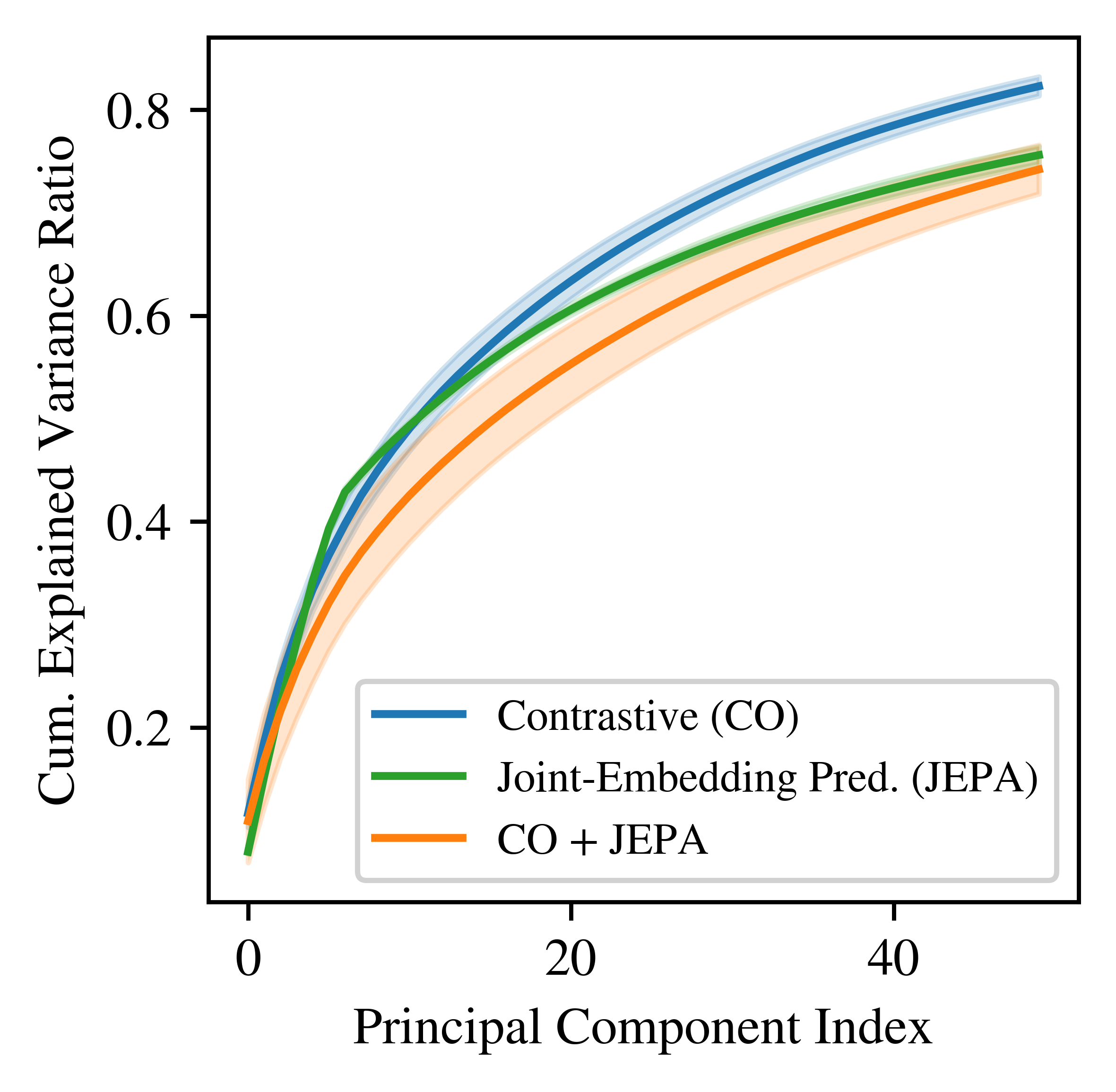}
			\caption{Cumulative Explained Variance}
			\label{fig:pca_varexpl}
		\end{subfigure}
	\end{minipage}
	\caption{Feature visualization: \textbf{(a)} Slices of the RGB encoded first three principal components of COJEPA. Brain structure contours in black for reference. Anatomical structures seem to be preserved in the principal components. \textbf{(b)} Mean cumulative explained variance for 10 random subjects and standard deviation of the 50 principial components for CO (blue), JEPA (green) and COJEPA (orange). COJEPA shows the most flat curve, indicating that it uses the space more efficiently than CO and JEPA.}
	\label{fig:featureexpl}
\end{figure}

\paragraph{Principal components of feature space.}
\cref{fig:featureexpl} visualizes the first three principal components of foreground patches as RGB channels. Similar areas of the brain are represented by similar colors. The edges of the brain structure as well as the ventricles are clearly depicted by red or pink colors. COJEPA embeddings (\cref{fig:pca_cojepa}) show a high dynamic range with clear color contrasts, indicating that anatomical structures are well-separated in the latent space. The cumulative explained variance (\Cref{fig:pca_varexpl}) confirms that COJEPA uses the latent space most efficiently: its spectrum is the flattest among the three methods, while CO and JEPA concentrate more variance in fewer dimensions.

\paragraph{Segmentation.}
\Cref{tab:res_seg} reports whole-tumor Dice on BraTS 2024 \citep{BRATS_2024} at four label fractions and after end-to-end finetuning. In the frozen probe regime, CO and COJEPA perform similarly across all label fractions, while JEPA lags notably at 5\% labels (Dice 0.43 vs.\ 0.52 for CO/COJEPA). After finetuning, all three methods converge to 0.70 to 0.71 Dice, demonstrating that the encoder features transfer well regardless of the pretraining objective. However, end-to-end finetuning increases Dice scores the most for JEPA (+0.08), while the benefit for CO and COJEPA is smaller (+0.04). Finetuning results close most of the gap to BrainIAC at 0.79 \citep{BRAINIAC}, despite BrainIAC being fine-tuned end-to-end on FLAIR sequences.

\begin{table}[!htb]
	\setlength{\tabcolsep}{4pt}
	\centering
	\caption{BraTS whole-tumor Dice ($\uparrow$): probing at 5/25/50/100\% training labels (N\@100\% = 1135) and end-to-end finetuning at 100\%.}
	\label{tab:res_seg}
	\begin{tabular}{lccccc}
		\toprule
		       & \multicolumn{4}{c}{frozen probe} & f-tune                                                    \\
		\cmidrule(lr){2-5}\cmidrule(l){6-6}
		Method & 5\%                              & 25\%         & 50\%         & 100\%        & 100\%        \\
		\midrule
		CO     & .52\spm{.01}                     & .61\spm{.01} & .64\spm{.01} & .67\spm{.01} & .71\spm{.01} \\
		JEPA   & .43\spm{.01}                     & .56\spm{.01} & .59\spm{.01} & .62\spm{.01} & .70\spm{.01} \\
		COJEPA & .52\spm{.01}                     & .61\spm{.01} & .64\spm{.01} & .66\spm{.01} & .71\spm{.01} \\
		\bottomrule
	\end{tabular}
\end{table}

\begin{figure}[t]
	\centering
	\includegraphics[width=.85\linewidth]{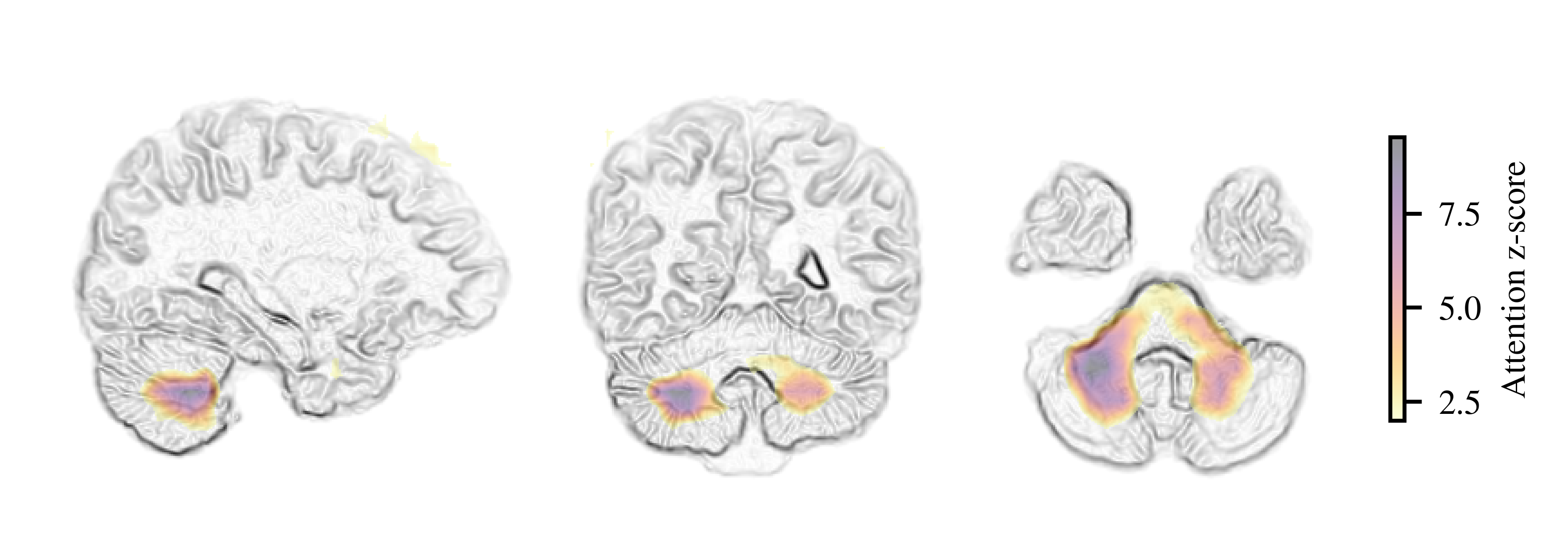}
	\label{fig:attn_age}
	\caption{Probe attention z-scores: Sliced heatmap of positive z-scores contributing to the linear prediction of age. Z-scores originate from the attention score of the query vector and each spatial token. The cerebellum appears to be the main source for age prediction.}
\end{figure}

\paragraph{Age prediction.}
\Cref{tab:res_age} reports age MAE on the OpenBHB external test set (3.0\,T, $N{=}339$). CO achieves the best frozen-probe MAE at 100\% labels (3.44\,yr using \cls{}), while COJEPA shows better data-efficiency at 5\% labels (4.22\,yr using \avg vs.\ 5.23\,yr using \sattn for CO) and the best finetuning MAE (2.55\,yr using \savg). COJEPA also shows more consistent performance across all pooling mechanisms compared to CO and JEPA. \Cref{fig:attn_age} shows a heatmap of the patches that contribute to age prediction, trilinearly interpolated to voxel space. The cerebellum appears to be the main source for age prediction. Per-site breakdowns are provided in \cref{tab:age_site_co,tab:age_site_jepa,tab:age_site_cojepa}.

\begin{table}[!htb]
	\setlength{\tabcolsep}{4pt}
	\centering
	\caption{OpenBHB age MAE ($\downarrow$, yr) on 3.0T scans: frozen probe at 5/25/50/100\% labels and end-to-end finetuning at 100\%.}
	\label{tab:res_age}
	\begin{tabular}{llccccc}
		\toprule
		                        &        & \multicolumn{4}{c}{frozen probe} & f-tune                                                                            \\
		\cmidrule(lr){3-6}\cmidrule(l){7-7}
		Method                  &        & 5\%                              & 25\%               & 50\%               & 100\%              & 100\%              \\
		\midrule
		\multirow{3}{*}{CO}     & \scls  & 5.85\spm{.25}                    & \textbf{3.59}\spm{.18} & \textbf{3.49}\spm{.17} & \textbf{3.44}\spm{.17} & 2.87\spm{.17}      \\
		                        & \savg  & 9.20\spm{.30}                    & 6.08\spm{.25}      & 6.86\spm{.24}      & 6.76\spm{.24}      & 2.90\spm{.16}      \\
		                        & \sattn & 5.23\spm{.23}                    & 4.06\spm{.19}      & 3.88\spm{.19}      & 3.83\spm{.19}      & 3.48\spm{.17}      \\
		\midrule[0.1pt]
		\multirow{2}{*}{JEPA}   & \savg  & 5.77\spm{.25}                    & 4.26\spm{.19}      & 4.33\spm{.19}      & 4.38\spm{.19}      & 3.04\spm{.17}      \\
		                        & \sattn & 8.37\spm{.29}                    & 5.04\spm{.22}      & 4.09\spm{.19}      & 4.12\spm{.19}      & 2.93\spm{.17}      \\
		\midrule[0.1pt]
		\multirow{3}{*}{COJEPA} & \scls  & 4.84\spm{.21}                    & 4.23\spm{.20}      & 4.21\spm{.20}      & 3.98\spm{.20}      & 2.64\spm{.17}      \\
		                        & \savg  & \textbf{4.22}\spm{.20}               & 3.78\spm{.19}      & 3.64\spm{.19}      & 3.60\spm{.18}      & \textbf{2.55}\spm{.17} \\
		                        & \sattn & 4.37\spm{.21}                    & 3.99\spm{.19}      & 3.93\spm{.19}      & 3.83\spm{.19}      & 2.73\spm{.17}      \\
		\bottomrule
	\end{tabular}
\end{table}

\begin{figure}[t]
	\centering
	\begin{subfigure}{0.494\textwidth}
		\centering
		\includegraphics[width=1\linewidth]{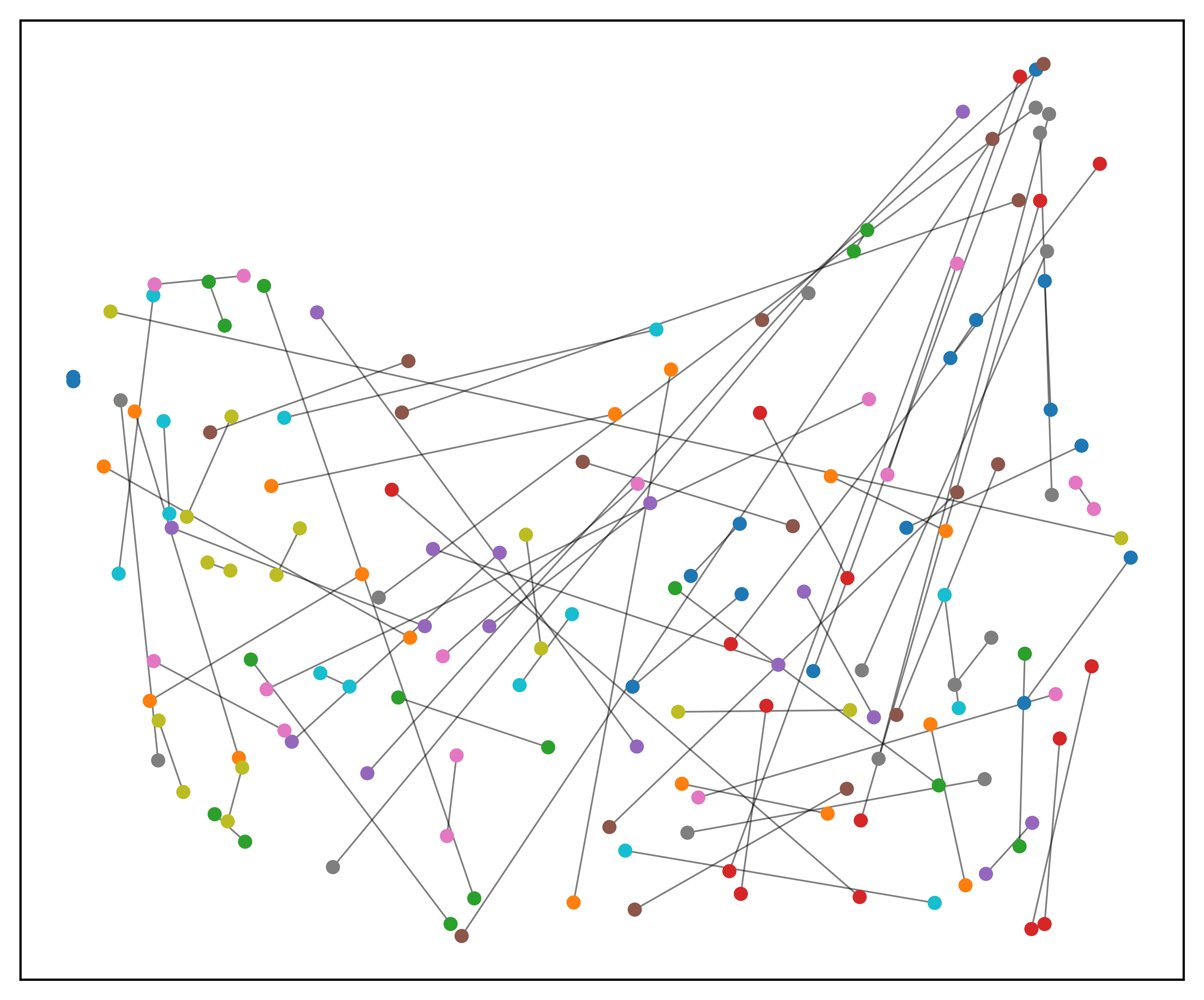}
		\caption{DZ Twins}
		\label{fig:umap_dz}
	\end{subfigure}
	\begin{subfigure}{0.494\textwidth}
		\centering
		\includegraphics[width=1\linewidth]{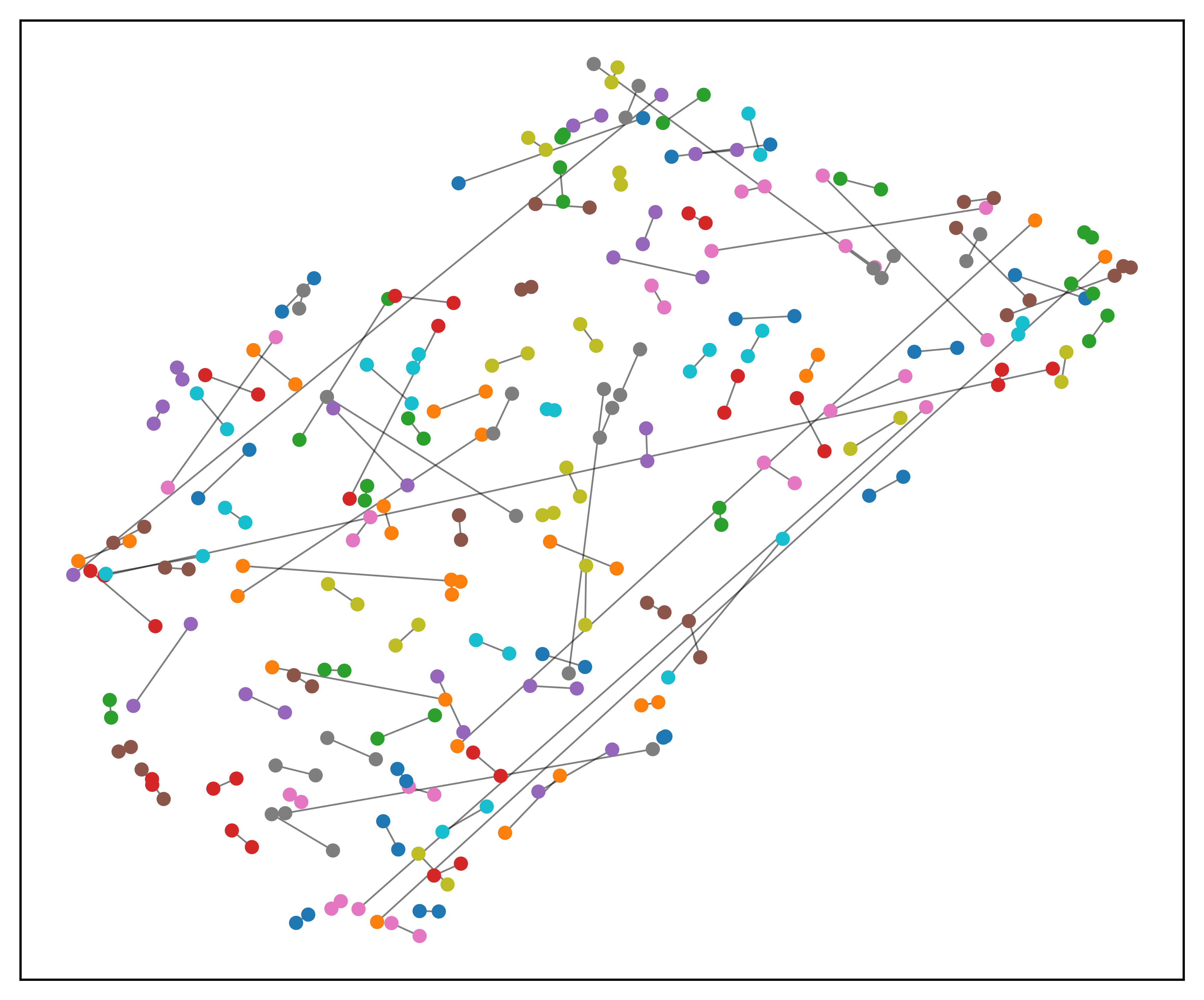}
		\caption{MZ Twins}
		\label{fig:umap_mz}
	\end{subfigure}
	\caption{UMAP plot of COJEPA \cls{} embeddings: \textbf{(a)} DZ twin pairs and \textbf{(b)} MZ twin pairs, connected via a line. Twin pairs are generally placed closer to each other than unrelated subjects.}
	\label{fig:umap}
\end{figure}

\paragraph{Retrieval.}
\Cref{tab:retrieval_metrics} reports zero-shot twin retrieval on the HCP hold-out. Pure JEPA fails at identity-style retrieval (MZ R@1\,=\,0.26), confirming that the predictive objective alone does not produce subject-discriminative global features. Both CO and COJEPA achieve strong MZ retrieval, with COJEPA reaching the best MZ R@1 (0.84). CO is competitive at higher ranks for DZ pairs. \Cref{fig:umap} shows a UMAP \citep{UMAP} plot of the COJEPA \cls{} embeddings; except for a few outliers, twin pairs are mapped closer to each other than unrelated subjects.

\begin{table}[!htb]
	\setlength{\tabcolsep}{4pt}
	\centering
	\caption{Zero-shot twin retrieval on the HCP hold-out (MZ $N{=}276$, DZ $N{=}158$).}
	\begin{adjustbox}{max width=\textwidth}
		\begin{tabular}{llcccccc}
			\toprule
			\multirow{2}[3]{*}{Method} &       & \multicolumn{3}{c}{MZ twins $\uparrow$} & \multicolumn{3}{c}{DZ twins $\uparrow$}                                                                                 \\
			\cmidrule(lr){3-5}\cmidrule(l){6-8}
			                           &       & R@1                                     & R@5                                     & R@10              & R@1               & R@5               & R@10              \\
			\midrule
			\multirow{2}{*}{CO}        & \scls & .78\spm{.03}                            & .93\spm{.02}                            & .96\spm{.01}      & .16\spm{.03}      & \textbf{.42}\spm{.04} & \textbf{.59}\spm{.04} \\
			                           & \savg & .76\spm{.03}                            & .90\spm{.02}                            & .94\spm{.01}      & .10\spm{.02}      & .31\spm{.04}      & .46\spm{.04}      \\
			\midrule[0.1pt]
			JEPA                       & \savg & .26\spm{.03}                            & .47\spm{.03}                            & .58\spm{.03}      & .07\spm{.02}      & .18\spm{.03}      & .29\spm{.04}      \\
			\midrule[0.1pt]
			\multirow{2}{*}{COJEPA}    & \scls & \textbf{.84}\spm{.02}                       & \textbf{.94}\spm{.02}                       & .96\spm{.01}      & \textbf{.17}\spm{.03} & .41\spm{.04}      & .54\spm{.04}      \\
			                           & \savg & .84\spm{.02}                            & \textbf{.94}\spm{.01}                       & \textbf{.97}\spm{.01} & \textbf{.17}\spm{.03} & .37\spm{.04}      & .54\spm{.04}      \\
			\bottomrule
		\end{tabular}
	\end{adjustbox}
	\label{tab:retrieval_metrics}
\end{table}

\section{Discussion}

The most striking result is the substantial improvement in zero-shot twin retrieval from the contrastive objective. JEPA, trained with a predictive objective alone, already encodes sufficient identity-relevant structure to retrieve MZ twins above chance (R@1\,=\,0.26), consistent with the well-established heritability of brain morphology~\citep{yao2023}. However, it largely fails on the more genetically distant DZ pairs (R@1\,=\,0.07), suggesting that the predictive objective alone does not produce sufficiently discriminative global representations. CO achieves strong MZ retrieval (R@1\,=\,0.78), and COJEPA further improves upon this (R@1\,=\,0.84), confirming that enforcing view-invariance within a subject produces global representations that generalize to genetic similarity.

A secondary finding is that global-average-pooled spatial tokens outperform the \cls token for retrieval, despite the contrastive loss being applied exclusively to the \cls token, suggesting that the contrastive signal propagates beneficially to the spatial representations through the attention layers.

The segmentation results show marginal but consistent differences across the three objectives, with the contrastive models performing slightly better on both tasks, indicating that the contrastive objective does not degrade and in fact slightly improves the local predictive quality of the representations. Similarly, age and sex prediction results are largely comparable across both models, with the exception of age prediction using the \cls probe, where performance drops when the contrastive loss is added. We speculate that the contrastive objective encourages the \cls token to focus on view-invariant subject identity at the expense of more continuous demographic variation such as age.
Notably, the attention-pooling age probe performs on par with dedicated brain age models trained on the same cohort \citep{sarica2024, kopetzky2024, he2022}, despite our encoder being trained in a fully self-supervised manner without any age-related supervision. This suggests that the learned representations capture morphological variation relevant to brain ageing as an emergent property of the pretraining objective

\section{Limitations}

Several limitations of this work should be acknowledged. The pretraining data of $N{=}2286$ healthy subjects is small by SSL standards, and the consistently higher error on the 1.5T site (site~36) suggests that scanner-domain gaps deserve dedicated study. Larger and more diverse pretraining cohorts would likely yield stronger and more generalizable representations. Additionally, due to the substantial computational cost of pretraining on volumetric MRI, an extensive hyperparameter search was not feasible, and the reported results may therefore not reflect the full potential of the proposed framework.

\section{Conclusion}

We presented COJEPA, a self-supervised framework for volumetric brain MRI that combines a joint-embedding predictive objective with a contrastive loss. The predictive objective produces spatially structured representations that support out-of-distribution segmentation and age prediction, as well as coarse identity retrieval. Specifically, the CO term adds discriminability needed for fine-grained subject-level retrieval. Trained on 2286 healthy subjects, the model achieves competitive downstream performance. Future work should investigate larger and more diverse pretraining cohorts, as well as the applicability of COJEPA representations to a broader range of clinical downstream tasks.


\section*{Acknowledgements}
This work was supported by the Danish Data Science Academy, which is funded by the Novo Nordisk Foundation (NNF21SA0069429) and VILLUM FONDEN (40516). The research was further supported by the Novo Nordisk Foundation grant NF22OC0076907 ”Cognitive spaces - Next generation explainability” and the Pioneer Centre for AI, DNRF grant number P1. Computational resources were provided by the DeiC National HPC (DeiC-DTU-N5-2024049).

%
%
\bibliography{main}
\bibliographystyle{tmlr}

\appendix

\crefalias{section}{appendix}

\section{Hyperparameters}

\begin{table}[h!]
	\centering
	\caption{Pretraining hyperparameters.}
	\label{tab:pretrain_shared}
	\begin{tabular}{ll}
		\toprule
		\textbf{Hyperparameter}     & \textbf{Value}        \\
		\midrule
		\multicolumn{2}{l}{\textit{Architecture}}           \\
		Model                       & ViT-Base              \\
		Patch size                  & $9^3$                 \\
		Patch embedding             & Conv + BatchNorm      \\
		Positional embedding        & Global, prepended     \\
		Register tokens             & 1                     \\
		Predictor depth             & 12                    \\
		Predictor embedding dim     & 384                   \\
		\midrule
		\multicolumn{2}{l}{\textit{Data}}                   \\
		Dataset                     & HCP Young Adult S1200 \\
		Input channels              & 1                     \\
		ROI size                    & $90^3$                \\
		Scale range                 & $[0.9,\ 1.5]$         \\
		Batch size (per node)       & 12                    \\
		\midrule
		\multicolumn{2}{l}{\textit{Masking}}                \\
		Context mask scale          & $[0.80,\ 1.00]$       \\
		Target mask scale           & $[0.20,\ 0.25]$       \\
		Target mask aspect ratio    & $[0.75,\ 1.25]$       \\
		Number of context masks     & 1                     \\
		Number of target masks      & 3                     \\
		Mask background tokens      & True                  \\
		\midrule
		\multicolumn{2}{l}{\textit{Optimization}}           \\
		Epochs                      & 4000                  \\
		Peak learning rate          & $1 \times 10^{-4}$    \\
		Start / final learning rate & $1 \times 10^{-5}$    \\
		Warmup epochs               & 50                    \\
		Weight decay (init / final) & $0.04\ /\ 0.40$       \\
		EMA momentum range          & $[0.996,\ 1.000]$     \\
		Mixed precision             & bfloat16              \\
		\midrule
		\multicolumn{2}{l}{\textit{Loss}}                   \\
		Prediction loss coefficient & 1.0                   \\
		Density loss coefficient    & 0.0                   \\
		Contrastive temperature     & 0.07                  \\
		Token norm                  & False                 \\
		\bottomrule
	\end{tabular}
\end{table}

\section{Ablations}

\Cref{tab:ablation} reports a 3D-component ablation of COJEPA on the HCP hold-out, measuring reconstruction quality (R²), representation rank (RankMe~\cite{RANKME}), and MZ twin retrieval at R@1/5/10. Removing global positional encodings causes the largest drop across all retrieval metrics, confirming that world-space coordinates are essential for learning subject-discriminative features. Removing the convolutional stem degrades retrieval moderately, while removing foreground masking has little effect on retrieval and slightly improves R².

\begin{table}[h]
	\centering
	\caption{3D-component ablation of COJEPA on the HCP hold-out.}
	\label{tab:ablation}
	\begin{adjustbox}{max width=\textwidth}
		\begin{tabular}{lcccccc}
			\toprule
			Variant                 & R$^2$          & RankMe         & R@1            & R@5            & R@10           \\
			\midrule
			Full (COJEPA)           & .572\spm{.003} & 753.6\spm{0.2} & .399\spm{.030} & .681\spm{.028} & .775\spm{.025} \\
			w/o global pos.\ enc.\  & .519\spm{.004} & 743.9\spm{0.5} & .127\spm{.020} & .283\spm{.027} & .377\spm{.029} \\
			w/o foreground mask     & .593\spm{.003} & 752.1\spm{0.2} & .424\spm{.030} & .703\spm{.028} & .768\spm{.025} \\
			w/o conv stem           & .567\spm{.003} & 752.2\spm{0.2} & .330\spm{.028} & .616\spm{.029} & .728\spm{.027} \\
			\bottomrule
		\end{tabular}
	\end{adjustbox}
\end{table}

\section{Age prediction}

\Cref{tab:age_site_co,tab:age_site_jepa,tab:age_site_cojepa} report age MAE broken down by acquisition site at all label fractions. Site~36 (1.5T) consistently shows the highest error across all methods, highlighting the domain gap between scanner field strengths. Sites 15 and 64, despite having few subjects ($N{\leq}20$), achieve competitive MAE on 3.0T scans.

\begin{table}[h]
	\small
	\centering
	\caption{Age MAE ($\downarrow$, yr) by acquisition site --- CO.}
	\label{tab:age_site_co}
	\begin{adjustbox}{max width=\textwidth}
		\begin{tabular}{ccc lcccc}
			\toprule
			Site                & T                    & $N$                  & Token           & 5\%             & 25\%            & 50\%            & 100\%           \\
			\midrule
			\multirow{3}{*}{15} & \multirow{3}{*}{3.0} & \multirow{3}{*}{20}
			                    & \sattn               & 3.84\spm{.70}        & 4.44\spm{.89}   & 5.19\spm{.95}   & 5.39\spm{.94}                                       \\
			                    &                      &                      & \savg           & 10.32\spm{1.29} & 10.28\spm{1.19} & 11.70\spm{1.17} & 11.04\spm{1.17} \\
			                    &                      &                      & \scls           & 4.44\spm{.65}   & 2.99\spm{.51}   & 2.80\spm{.53}   & 3.15\spm{.52}   \\
			\midrule
			\multirow{3}{*}{19} & \multirow{3}{*}{3.0} & \multirow{3}{*}{32}
			                    & \sattn               & 5.61\spm{.80}        & 4.81\spm{.88}   & 4.79\spm{.86}   & 4.68\spm{.88}                                       \\
			                    &                      &                      & \savg           & 9.87\spm{.88}   & 7.57\spm{.69}   & 8.00\spm{.67}   & 7.07\spm{.67}   \\
			                    &                      &                      & \scls           & 7.36\spm{1.23}  & 5.29\spm{.93}   & 4.92\spm{.82}   & 4.81\spm{.83}   \\
			\midrule
			\multirow{3}{*}{36} & \multirow{3}{*}{1.5} & \multirow{3}{*}{56}
			                    & \sattn               & 9.41\spm{.99}        & 12.68\spm{1.16} & 13.87\spm{1.25} & 14.21\spm{1.22}                                     \\
			                    &                      &                      & \savg           & 8.24\spm{.81}   & 9.54\spm{1.15}  & 10.22\spm{1.21} & 10.87\spm{1.23} \\
			                    &                      &                      & \scls           & 9.00\spm{.86}   & 13.78\spm{1.30} & 14.59\spm{1.29} & 13.61\spm{1.23} \\
			\midrule
			\multirow{3}{*}{41} & \multirow{3}{*}{3.0} & \multirow{3}{*}{103}
			                    & \sattn               & 6.18\spm{.44}        & 3.43\spm{.34}   & 3.29\spm{.31}   & 3.42\spm{.32}                                       \\
			                    &                      &                      & \savg           & 12.62\spm{.52}  & 7.04\spm{.46}   & 7.26\spm{.43}   & 7.73\spm{.43}   \\
			                    &                      &                      & \scls           & 6.00\spm{.46}   & 2.96\spm{.27}   & 3.07\spm{.29}   & 3.05\spm{.28}   \\
			\midrule
			\multirow{3}{*}{55} & \multirow{3}{*}{3.0} & \multirow{3}{*}{166}
			                    & \sattn               & 4.90\spm{.32}        & 4.18\spm{.24}   & 3.83\spm{.23}   & 3.66\spm{.23}                                       \\
			                    &                      &                      & \savg           & 6.95\spm{.36}   & 4.45\spm{.29}   & 5.59\spm{.31}   & 5.47\spm{.30}   \\
			                    &                      &                      & \scls           & 5.64\spm{.32}   & 3.70\spm{.23}   & 3.59\spm{.23}   & 3.44\spm{.23}   \\
			\midrule
			\multirow{3}{*}{64} & \multirow{3}{*}{3.0} & \multirow{3}{*}{18}
			                    & \sattn               & 3.63\spm{.72}        & 4.77\spm{.81}   & 4.58\spm{.79}   & 4.36\spm{.70}                                       \\
			                    &                      &                      & \savg           & 7.93\spm{1.03}  & 8.33\spm{1.12}  & 8.94\spm{1.12}  & 7.87\spm{1.06}  \\
			                    &                      &                      & \scls           & 5.74\spm{1.03}  & 3.84\spm{.76}   & 3.17\spm{.73}   & 3.58\spm{.71}   \\
			\bottomrule
		\end{tabular}
	\end{adjustbox}
\end{table}

\begin{table}[h]
	\small
	\centering
	\caption{Age MAE ($\downarrow$, yr) by acquisition site --- JEPA (\scls{} not available).}
	\label{tab:age_site_jepa}
	\begin{adjustbox}{max width=\textwidth}
		\begin{tabular}{ccc lcccc}
			\toprule
			Site                & T                    & $N$                  & Token           & 5\%             & 25\%            & 50\%            & 100\%           \\
			\midrule
			\multirow{2}{*}{15} & \multirow{2}{*}{3.0} & \multirow{2}{*}{20}
			                    & \sattn               & 3.96\spm{.53}        & 3.97\spm{.72}   & 3.50\spm{.63}   & 3.22\spm{.53}                                       \\
			                    &                      &                      & \savg           & 3.53\spm{.66}   & 4.91\spm{.70}   & 4.83\spm{.68}   & 4.33\spm{.54}   \\
			\midrule
			\multirow{2}{*}{19} & \multirow{2}{*}{3.0} & \multirow{2}{*}{32}
			                    & \sattn               & 6.17\spm{.88}        & 5.93\spm{.92}   & 5.31\spm{.87}   & 5.32\spm{.80}                                       \\
			                    &                      &                      & \savg           & 6.41\spm{.86}   & 5.22\spm{.79}   & 5.19\spm{.80}   & 5.21\spm{.80}   \\
			\midrule
			\multirow{2}{*}{36} & \multirow{2}{*}{1.5} & \multirow{2}{*}{56}
			                    & \sattn               & 15.88\spm{1.59}      & 18.52\spm{1.51} & 17.77\spm{1.56} & 17.51\spm{1.51}                                     \\
			                    &                      &                      & \savg           & 14.37\spm{1.56} & 16.27\spm{1.66} & 16.74\spm{1.67} & 16.12\spm{1.61} \\
			\midrule
			\multirow{2}{*}{41} & \multirow{2}{*}{3.0} & \multirow{2}{*}{103}
			                    & \sattn               & 8.21\spm{.43}        & 3.43\spm{.27}   & 3.16\spm{.26}   & 3.42\spm{.28}                                       \\
			                    &                      &                      & \savg           & 4.32\spm{.36}   & 3.37\spm{.28}   & 3.63\spm{.31}   & 4.12\spm{.34}   \\
			\midrule
			\multirow{2}{*}{55} & \multirow{2}{*}{3.0} & \multirow{2}{*}{166}
			                    & \sattn               & 9.70\spm{.44}        & 6.05\spm{.32}   & 4.45\spm{.28}   & 4.41\spm{.28}                                       \\
			                    &                      &                      & \savg           & 6.82\spm{.37}   & 4.40\spm{.26}   & 4.30\spm{.25}   & 4.26\spm{.26}   \\
			\midrule
			\multirow{2}{*}{64} & \multirow{2}{*}{3.0} & \multirow{2}{*}{18}
			                    & \sattn               & 5.73\spm{1.11}       & 4.47\spm{.71}   & 4.55\spm{.79}   & 4.39\spm{.74}                                       \\
			                    &                      &                      & \savg           & 5.89\spm{1.34}  & 5.58\spm{1.06}  & 6.50\spm{1.10}  & 5.50\spm{1.08}  \\
			\bottomrule
		\end{tabular}
	\end{adjustbox}
\end{table}

\begin{table}[h]
	\small
	\centering
	\caption{Age MAE ($\downarrow$, yr) by acquisition site --- COJEPA.}
	\label{tab:age_site_cojepa}
	\begin{adjustbox}{max width=\textwidth}
		\begin{tabular}{ccc lcccc}
			\toprule
			Site                & T                    & $N$                  & Token           & 5\%             & 25\%            & 50\%            & 100\%           \\
			\midrule
			\multirow{3}{*}{15} & \multirow{3}{*}{3.0} & \multirow{3}{*}{20}
			                    & \sattn               & 4.69\spm{.65}        & 3.84\spm{.66}   & 3.61\spm{.62}   & 3.47\spm{.64}                                       \\
			                    &                      &                      & \savg           & 4.48\spm{.55}   & 3.68\spm{.70}   & 3.97\spm{.66}   & 3.87\spm{.71}   \\
			                    &                      &                      & \scls           & 4.79\spm{.61}   & 4.63\spm{.91}   & 4.16\spm{.70}   & 4.32\spm{.72}   \\
			\midrule
			\multirow{3}{*}{19} & \multirow{3}{*}{3.0} & \multirow{3}{*}{32}
			                    & \sattn               & 5.08\spm{.64}        & 4.58\spm{.86}   & 4.77\spm{.83}   & 4.75\spm{.88}                                       \\
			                    &                      &                      & \savg           & 4.99\spm{.66}   & 4.51\spm{.87}   & 4.32\spm{.83}   & 4.25\spm{.87}   \\
			                    &                      &                      & \scls           & 4.96\spm{.71}   & 5.27\spm{.85}   & 5.51\spm{.87}   & 5.18\spm{.91}   \\
			\midrule
			\multirow{3}{*}{36} & \multirow{3}{*}{1.5} & \multirow{3}{*}{56}
			                    & \sattn               & 12.54\spm{1.22}      & 10.90\spm{1.17} & 13.92\spm{1.30} & 12.56\spm{1.22}                                     \\
			                    &                      &                      & \savg           & 11.92\spm{1.16} & 11.28\spm{1.19} & 13.12\spm{1.28} & 12.55\spm{1.23} \\
			                    &                      &                      & \scls           & 11.32\spm{1.06} & 12.15\spm{1.28} & 14.52\spm{1.40} & 13.56\spm{1.34} \\
			\midrule
			\multirow{3}{*}{41} & \multirow{3}{*}{3.0} & \multirow{3}{*}{103}
			                    & \sattn               & 3.58\spm{.36}        & 2.53\spm{.29}   & 2.58\spm{.32}   & 2.55\spm{.33}                                       \\
			                    &                      &                      & \savg           & 3.65\spm{.34}   & 2.78\spm{.29}   & 2.68\spm{.31}   & 3.16\spm{.32}   \\
			                    &                      &                      & \scls           & 4.41\spm{.38}   & 3.37\spm{.32}   & 3.12\spm{.34}   & 3.04\spm{.33}   \\
			\midrule
			\multirow{3}{*}{55} & \multirow{3}{*}{3.0} & \multirow{3}{*}{166}
			                    & \sattn               & 4.66\spm{.31}        & 4.66\spm{.26}   & 4.54\spm{.26}   & 4.33\spm{.25}                                       \\
			                    &                      &                      & \savg           & 4.31\spm{.30}   & 4.12\spm{.25}   & 3.92\spm{.25}   & 3.52\spm{.23}   \\
			                    &                      &                      & \scls           & 4.95\spm{.31}   & 4.31\spm{.25}   & 4.58\spm{.26}   & 4.16\spm{.25}   \\
			\midrule
			\multirow{3}{*}{64} & \multirow{3}{*}{3.0} & \multirow{3}{*}{18}
			                    & \sattn               & 4.57\spm{1.18}       & 5.33\spm{.88}   & 4.94\spm{.75}   & 5.27\spm{.87}                                       \\
			                    &                      &                      & \savg           & 5.06\spm{1.25}  & 5.11\spm{.93}   & 5.05\spm{.79}   & 5.43\spm{.91}   \\
			                    &                      &                      & \scls           & 6.08\spm{1.08}  & 6.10\spm{1.19}  & 4.79\spm{.92}   & 5.13\spm{.93}   \\
			\bottomrule
		\end{tabular}
	\end{adjustbox}
\end{table}

\clearpage



\end{document}